\documentclass{article}

\usepackage[preprint]{neurips_2026}

\makeatletter
\renewcommand{\@noticestring}{%
  Preprint. Submitted to the 9th Workshop on Machine Learning and the
  Physical Sciences (ML4PS 2026) at NeurIPS 2026.%
}
\makeatother

\usepackage[utf8]{inputenc}
\usepackage[T1]{fontenc}
\usepackage{hyperref}
\usepackage{url}
\usepackage{booktabs}
\usepackage{amsmath}
\usepackage{amsfonts}
\usepackage{amssymb}
\usepackage{bm}
\usepackage{graphicx}
\usepackage{subcaption}
\usepackage{nicefrac}
\usepackage{microtype}
\usepackage{xcolor}
\usepackage{wrapfig}

\newcommand{\vx}{\mathbf{x}}
\newcommand{\vy}{\mathbf{y}}
\newcommand{\vu}{\mathbf{u}}

\newcommand{\E}{\mathbb{E}}
\newcommand{\te}{T_e}

\title{Probabilistic and Geometry-Aware Neural Surrogate of
       Scrape-Off Layer Plasma Simulations}

\author{%
  Gabriele Gianuzzo \\
  Eindhoven University of Technology \\
  \texttt{g.gianuzzo@student.tue.nl} \\
  \And
  Stefan Dasbach \\
  DIFFER \\
  \texttt{S.F.W.Dasbach@differ.nl} \\
  \And
  Fleur Hendriks \\
  DIFFER \\
  \texttt{F.Hendriks@differ.nl} \\
  \AND
  Sven Wiesen \\
  DIFFER \\
  \texttt{S.Wiesen@differ.nl} \\
  \And
  Vlado Menkovski \\
  Eindhoven University of Technology \\
  \texttt{V.Menkovski@tue.nl} \\
}

\begin{document}

\maketitle

\begin{abstract}
Fast surrogates for tokamak boundary-plasma simulation are typically deterministic regressors mapping a global operating point to a flattened vector of cell values. Near the divertor detachment transition the steady state is not reliably single-valued. A point estimate must average over qualitatively different plasma states, and it arrives with no statement of confidence. Moreover, the flattened vector representation discards the geometric structure of the SOLPS-ITER mesh. This work addresses both problems. We unroll the curvilinear mesh into three fixed-size image tensors whose layout preserves cell adjacency and inverts exactly, letting a convolutional network act on the geometry without loss of information. A conditional flow-matching model, well suited to highly sensitive systems, is then trained on this representation. The result is an efficient, scalable surrogate that captures multiple plausible outcomes even at sensitive operating points. Along a gas-puff scan, the predictive distribution splits into a hot and a cold mode across an early regime transition. A further check on synthetic data with an injected bifurcation of known size confirms the model recovers both branches rather than their average.
\end{abstract}

\section{Introduction}
\label{sec:intro}

Power leaving the confined core of a tokamak crosses into the scrape-off layer (SOL), a narrow boundary region of open field lines, and is guided onto a small set of divertor target plates. Holding the resulting heat flux within what materials can survive is a gating constraint on next generation devices like ITER and DEMO \citep{stangeby2000, pitcher1997}. The main lever is detachment: heavier fueling, or seeding an impurity such as nitrogen or neon, moves exhaust power into radiation and into a neutral cloud that shields the plate, and the target then cools by orders of magnitude. Locating that transition requires a coupled plasma and neutral transport solve, for which the SOLPS-ITER \citep{bonnin2016} simulator is standard, at hours to weeks per converged steady state. Workloads needing hundreds of solves, such as uncertainty propagation or design search, are therefore tedious \citep{wiesen2024datadriven}. 

This motivates the use of a fast surrogate, trained on a dataset of simulations that can be queried at a negligible cost. Existing surrogates for this dataset map the eight global scalars to the fields using per-quantity multilayer perceptrons on flattened cell vectors \citep{dasbach2023surrogate, dasbach2026solpsnn}. Near detachment, very similar inputs admit qualitatively different steady states. Such bifurcating behaviour is hard to capture with a deterministic machine learning model because it requires a high level of precision and large amount of training data near the critical point. To overcome this, in this work we developed a probabilistic model based on conditional flow-matching that allows us to sample possible steady states for a given input vector and avoid the solution averaging that comes from deterministic models. A separate problem is geometric: flattening the mesh into a vector discards its spatial layout entirely. We instead unroll the curvilinear mesh into an adjacency-preserving image representation that keeps the convolutional inductive bias while fixing the geometry. We contribute a lossless adjacency-preserving encoding of the edge mesh as three CNN-ready tensors, a conditional flow-matching surrogate trained on it, and evidence that the resulting uncertainty is physically meaningful.

\section{Method}

\begin{wrapfigure}{r}{0.32\textwidth}
  \centering
  \includegraphics[width=\linewidth]{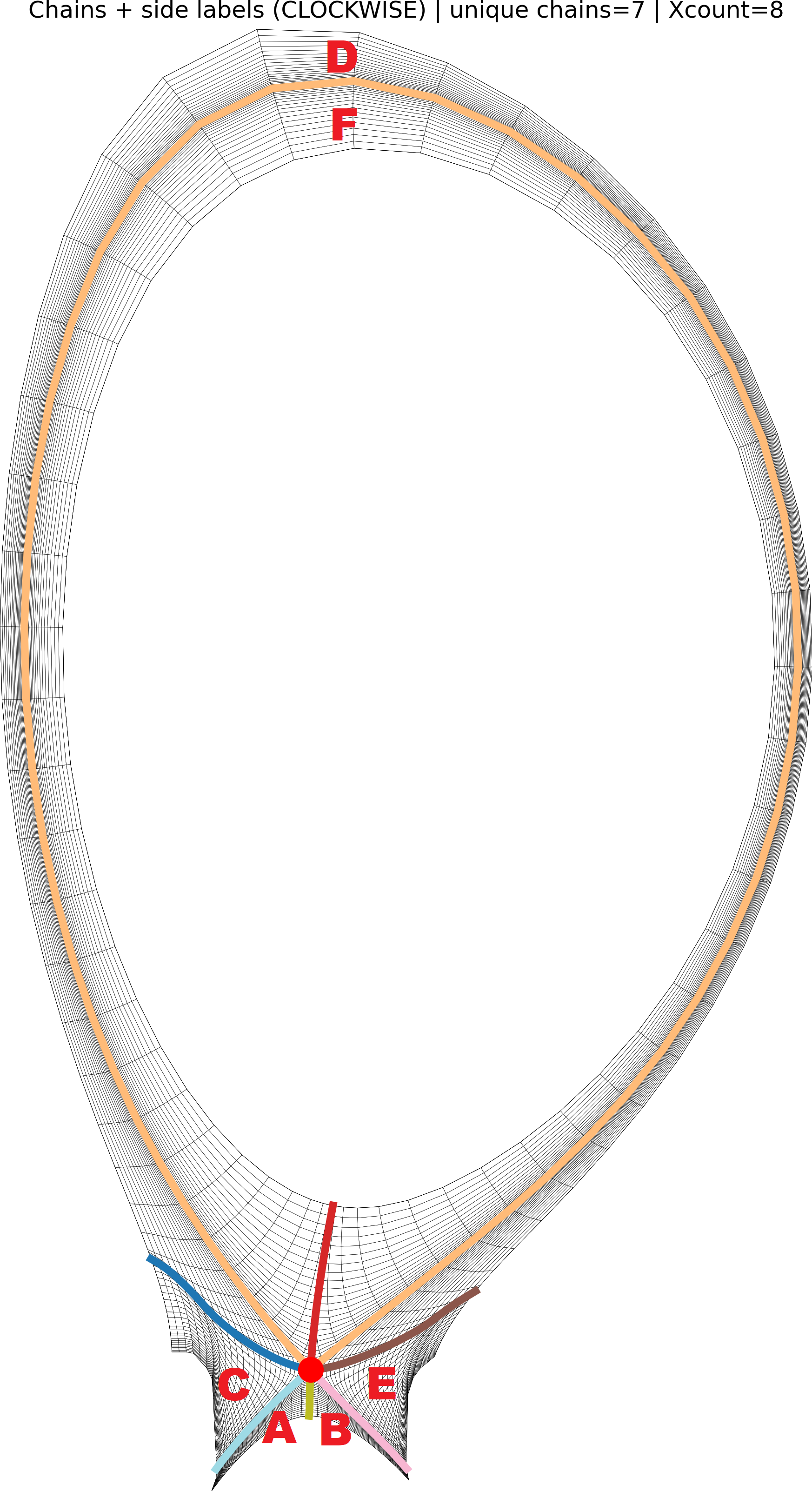}\\[3pt]
  {\footnotesize (a) The six strips on the native mesh, A--F.}\\[8pt]
  \includegraphics[width=\linewidth]{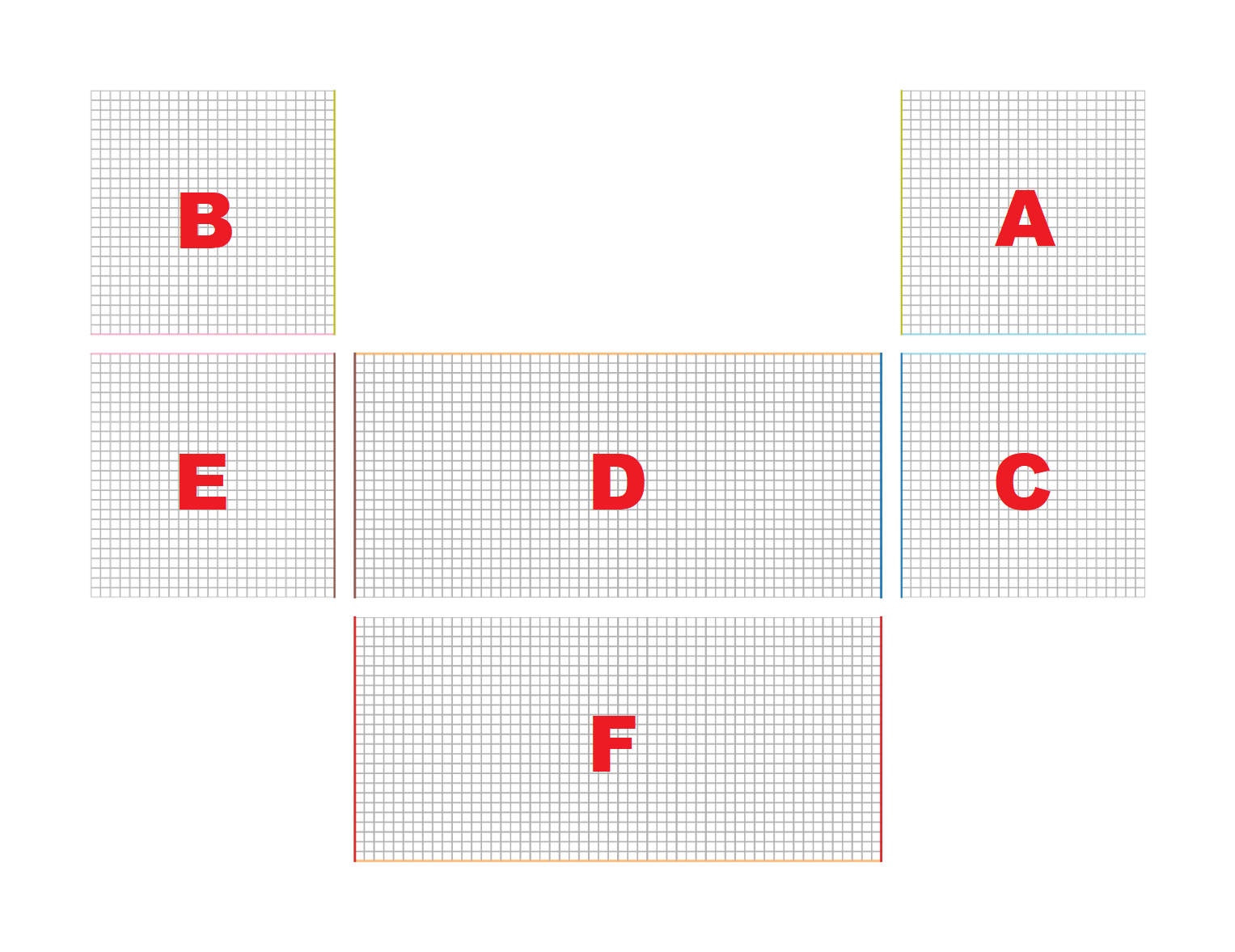}\\[3pt]
  {\footnotesize (b) Three-view layout before padding.}
  \caption{Matching border colours mark the seven adjacencies, each restored by padding and reinjected boundary.}
  \label{fig:views}
\end{wrapfigure}

\paragraph{Data.}
We use the SOLPS-ITER database of \citet{dasbach2023surrogate}: $7221$ converged steady states ($5756$ train, $1465$ test) on one fixed-topology $104 \times 50$ mesh. No new simulations were produced. 
Each run is summarised by eight global scalars (device size and field, input power, three particle-source rates, two transport coefficients) and yields $22$ field channels: two temperatures, ten species densities, ten parallel velocities. These span over ten orders of magnitude, so positive channels are mapped through $\log_{10}$ and signed ones through a symmetric logarithm, then standardised per channel over active pixels.

\paragraph{Geometry-aware three-view representation.}
A convolution presumes a filter is meaningful wherever it is applied, which
requires the pixel grid to reflect physical neighbourhood. Treating logical
mesh indices as pixel coordinates does not: cells adjacent in index space can
lie far apart in the poloidal plane, particularly near the X-point. We
partition the mesh into six connected strips, A to F (Figure~\ref{fig:views}a),
and unroll each onto a rectangular pixel block by traversing it clockwise in
an adjacency-preserving order, storing the cell-to-pixel assignment as an
explicit index map. Seven adjacencies survive between strips: $A\!-\!B$ and
$C\!-\!D\!-\!E$ within groups, $A\!-\!C$, $B\!-\!E$ and $F\!-\!D$ across them,
and a periodic wrap joining the ends of F.

No single canvas can place all six strips so that every one of the seven
adjacencies faces a matching gap (Figure~\ref{fig:views}b): pinning
$F\!-\!D$ centrally forces $A\!-\!C$ and $B\!-\!E$ to pull A and B to
opposite ends of the row, so $A\!-\!B$ is left without a facing edge and two
false seams appear instead. We therefore split the strips across three
canvases, top (A, B), mid (C, D, E) and bottom (F). Every one of the seven
adjacencies, whether within a canvas or across two, is restored the same
way: once at construction, by copying a narrow boundary band into the
padding gap facing each neighbour, cyclically for the wrap of F; and again
inside the network, by repeating that copy on feature maps after each block.
Reinjection keeps every reconnection alive at every scale, couples the three
views in one forward pass, and adds no parameters.

Views are embedded in a common $55 \times 134$ canvas for batching. Because
each active cell occupies one pixel of one view, and both directions are pure
index operations, the mapping is a bijection: all $5200$ cells are covered
exactly once, and the round trip is exact. Evaluation is therefore carried
out on the reconstructed mesh, without charging any error to the
representation itself.

\paragraph{Conditional flow matching over fields.}
We model $p(\vy^{(v)} \mid \vx, v)$ per view rather than a point estimate, with
one network shared across views and conditioned on the view label $v$.
Conditional flow matching learns a velocity field transporting a simple prior to
the data along a prescribed path \citep{lipman2023flow, liu2023rectified}. We use
the straight-line path: for a data field $\vy_1$ and an independent prior draw
$\vy_0 \sim \mathcal{N}(0, I)$ on the view's active pixels,
$\vy_t = (1-t)\vy_0 + t\vy_1$ and $\vu_t = \vy_1 - \vy_0$. Rather than regressing
$\vu_t$ we use the equivalent data-prediction parameterisation
\citep{lipman2024fmguide}, training the network output
$\hat{\vy}_1(\vx, \vy_t, t, v, \vy_{\mathrm{sc}})$, where $\vy_{\mathrm{sc}}$ is a
self-conditioning input \citep{chen2023analogbits}, under a mask-weighted mean
absolute error and recovering the velocity in closed form,
\begin{equation}
  \mathcal{L} = \E_{\vx, \vy_1, \vy_0, t}
  \big\| \hat{\vy}_1 - \vy_1 \big\|_{\mathcal{M}} ,
  \qquad
  v_\theta = (\hat{\vy}_1 - \vy_t)/(1 - t) ,
  \label{eq:loss}
\end{equation}
with the denominator clamped at $10^{-3}$; the loss is averaged equally over
channels and views. During training, times are drawn uniformly on $[0.01, 0.99]$, one per simulation, shared across its views. At inference, sampling integrates the probability-flow ODE with $K$ Euler steps, the views advancing in lockstep. Over $S$ prior draws the per-pixel sample mean is the point prediction and the
per-pixel standard deviation the uncertainty, both on physical fields after
the transforms are inverted.

\paragraph{Network and training.}
The predictor is a U-Net \citep{ronneberger2015unet} with two encoders. A condition encoder reads only the conditioning channels (mask, the eight
scalars broadcast over the active region, cell centroid coordinates, a learned view
embedding), giving the decoder features that stay constant along a trajectory;
a main encoder reads those channels concatenated with $\vy_t$ and the
self-conditioning field. The decoder fuses both at each resolution and is
modulated by a sinusoidal embedding of $t$ through FiLM \citep{perez2018film};
the dedicated conditioning route follows ControlNet
\citep{zhang2023controlnet}. At base width $64$ the model has $9.4$M
parameters. We train with AdamW \citep{loshchilov2019adamw}, learning rate
$3\times10^{-4}$, weight decay $10^{-4}$, cosine annealed, batch size $16$, for
$200$ epochs, reporting the best-validation checkpoint of a $90/10$ split.

\begin{table}[htbp]
  \centering
  \caption{Electron-temperature fidelity by regime, computed from $S=50$
prior draws per operating point over the full test set. Log MAE is the
per-pixel mean $|\log_{10}(\mathrm{pred}) - \log_{10}(\mathrm{truth})|$, reported with its standard error; log bias is the median of $\log_{10}(\mathrm{pred}) - \log_{10}(\mathrm{truth})$.}
  \label{tab:fidelity}
  \begin{tabular}{lcccccc}
    \toprule
     & \multicolumn{2}{c}{2D interior} & \multicolumn{2}{c}{OMP profile} & \multicolumn{2}{c}{OT profile} \\
    \cmidrule(lr){2-3} \cmidrule(lr){4-5} \cmidrule(lr){6-7}
    Regime & log MAE & log bias & log MAE & log bias & log MAE & log bias \\
    \midrule
    Sheath-limited & $0.132 \pm 0.010$ & $-0.025$ & $0.108 \pm 0.009$ & $-0.018$ & $0.158 \pm 0.011$ & $-0.031$ \\
    Attached       & $0.093 \pm 0.009$ & $-0.004$ & $0.055 \pm 0.007$ & $-0.002$ & $0.149 \pm 0.010$ & $-0.003$ \\
    Detached       & $0.197 \pm 0.015$ & $-0.019$ & $0.148 \pm 0.015$ & $-0.015$ & $0.203 \pm 0.013$ & $-0.021$ \\
    Cold-core      & $0.614 \pm 0.041$ & $+0.343$ & $0.654 \pm 0.044$ & $+0.427$ & $0.776 \pm 0.051$ & $+0.507$ \\
    \midrule
    All            & $0.201 \pm 0.009$ & $-0.005$ & $0.176 \pm 0.009$ & $-0.003$ & $0.254 \pm 0.011$ & $-0.004$ \\
    \bottomrule
  \end{tabular}
\end{table}

\section{Results}
\label{sec:results}

All results use $K = 50$ Euler steps throughout, and $S = 50$ prior draws unless noted otherwise. The gas-puff scan's distributional analysis and the synthetic verification use $S = 500$ prior draws, for finer resolution of the predictive distribution at the
smaller number of points examined there. Results report the electron temperature $\te$, which sets the divertor regime and governs detachment. Regimes follow \citet{dasbach2026solpsnn}, defined from $\te$ at the outer midplane (OMP) and outer target (OT): cold-core, detached, attached, and sheath-limited, in that order of precedence. Metrics are per regime, a global average being dominated by the most populous one, and fidelity is measured in $\log_{10}$ space, which does not degenerate on the near-zero cold-core field. A fifty-sample estimate for one operating point costs $79.4$~s ($1.59$~s per sampled field), against hours to weeks for a single SOLPS-ITER solve.

\paragraph{Reconstruction fidelity.}
Reading sample spread as uncertainty is meaningful only if the predictive mean is
faithful. Across the three regimes carrying the divertor physics the 2D field is
reconstructed with negligible bias to a multiplicative error of a factor
$1.2$--$1.6$ (Table~\ref{tab:fidelity}). The exception is cold-core, where the model
over-predicts a field that is itself near zero by a median factor of about two in two dimensions and three at the target,
though the absolute error stays below one electronvolt (median $0.44$~eV).
Errors are set by a heavy tail, the twenty worst of $1465$ test cases
accounting for over half the global mean absolute error.

\begin{figure}[htbp]
  \centering
  \begin{subfigure}[b]{0.33\textwidth}
    \includegraphics[width=\linewidth]{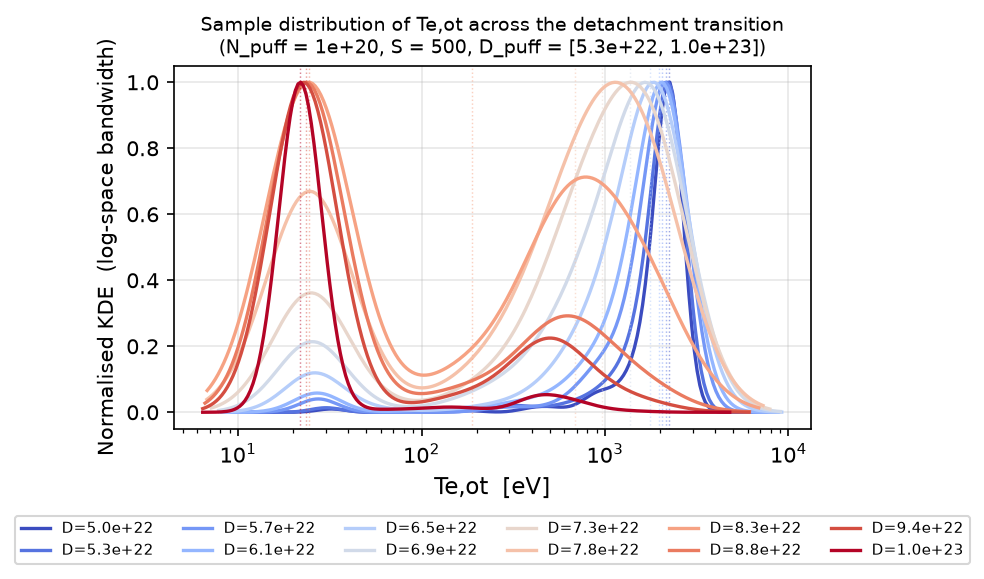}
    \caption{Model samples, gas-puff scan.}
  \end{subfigure}
  \hfill
  \begin{subfigure}[b]{0.32\textwidth}
    \includegraphics[width=\linewidth]{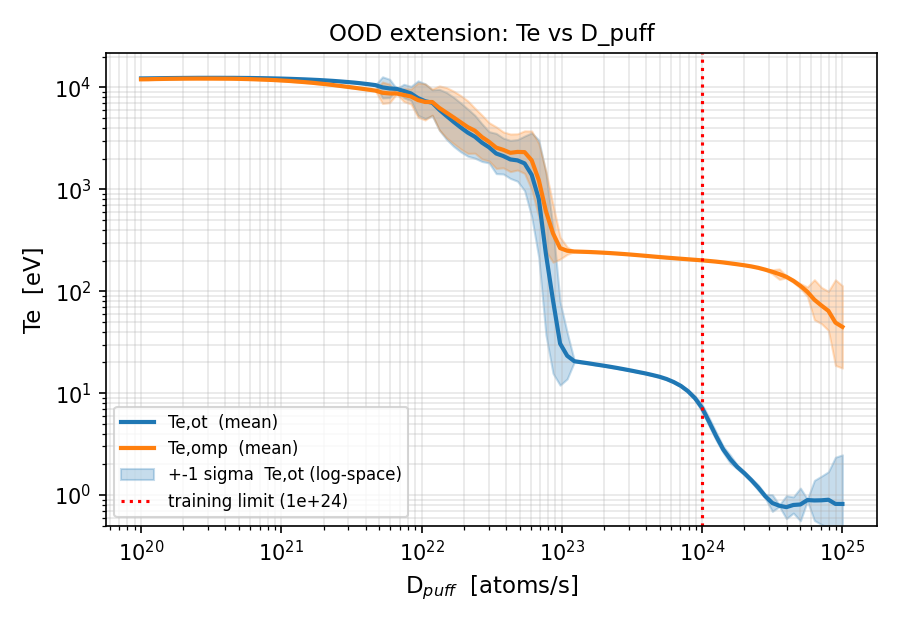}
    \caption{OOD extension, same scan.}
  \end{subfigure}
  \hfill
  \begin{subfigure}[b]{0.33\textwidth}
    \includegraphics[width=\linewidth]{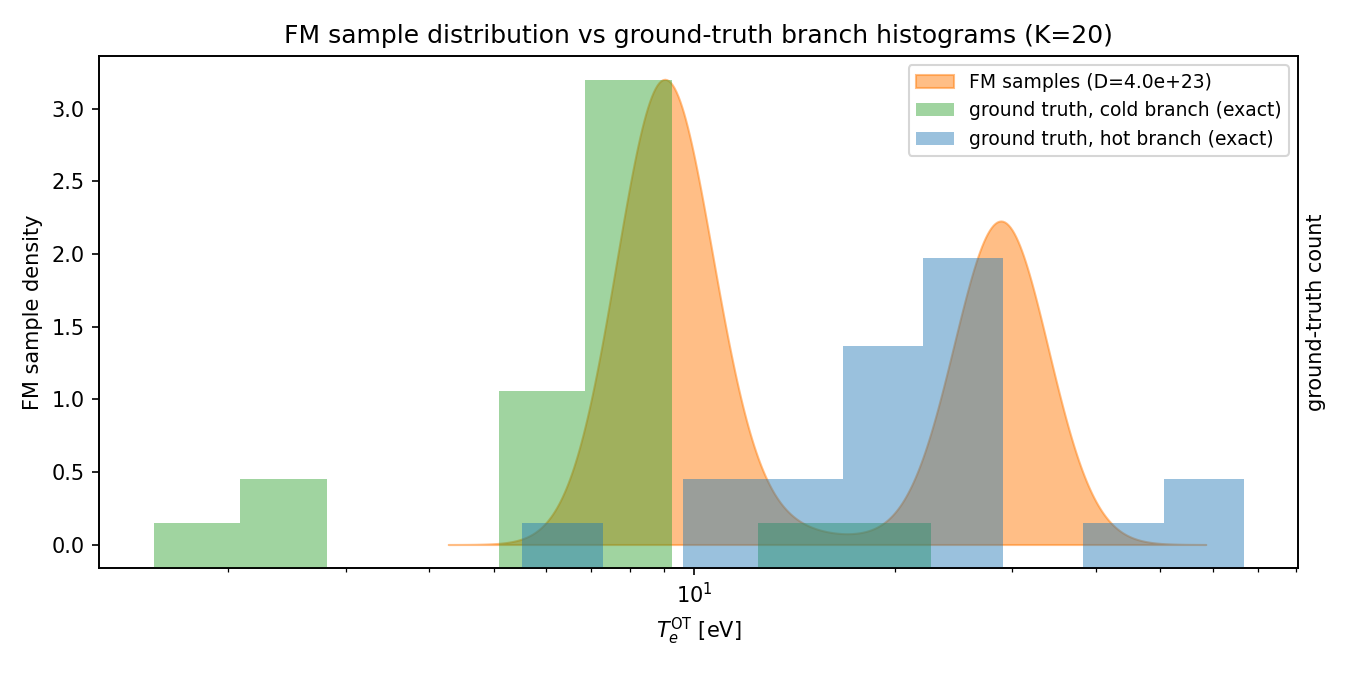}
    \caption{Synthetic data: single-input FM samples vs.\ pooled ground truth.}
  \end{subfigure}
  \caption{(a) Sampled $\te^{\mathrm{OT}}$ along the gas-puff slice: one tight
cluster on either branch, two populations across the transition. (b) Mean
predicted $\te^{\mathrm{OT}}$ and $\te^{\mathrm{OMP}}$ one decade past the
training limit (dotted line), $\pm 1\sigma$ on $\te^{\mathrm{OT}}$ in log space: the band widens again near the extrapolation limit (c) The synthetic-trained model past the transition
($D_{\mathrm{puff}} = 4\times10^{23}\,\mathrm{s^{-1}}$) against exact branch
temperatures from the $20$ nearest test points; both are recovered.}
  \label{fig:results}
\end{figure}

\paragraph{Uncertainty across a gas-puff scan.}
The predicted mean $\te^{\mathrm{OT}}$ reproduces the gas-puff response of \citet{dasbach2026solpsnn}. Along a slice at $N_{\mathrm{puff}} =
10^{20}\,\mathrm{s^{-1}}$, $\te^{\mathrm{OT}}$ falls by more than two orders
of magnitude between the sheath-limited boundary near $D_{\mathrm{puff}}
\approx 5\times10^{22}\,\mathrm{s^{-1}}$ and the training limit at
$10^{24}\,\mathrm{s^{-1}}$, while $\te^{\mathrm{OMP}}$ falls by about one
order and plateaus near $200$~eV: the target cools far faster than upstream,
the signature of scrape-off layer power dissipation (Figure~\ref{fig:results}b).

The uncertainty along the sweep is not uniform. Relative spread climbs from a few percent on either branch to of order
$200\%$ across the transition window, and it is a split rather than a diffuse
band: inside the window the draws separate into a hot and a cold population with
few between (Figure~\ref{fig:results}a), whereas on either branch they form one
tight cluster. That the ambiguity belongs to the data is shown by the raw simulations: the
closest input pair among the $7221$ runs, with puff rates differing by under
$0.2\%$, has $\te^{\mathrm{OT}} = 300$~eV in one run and $9.9$~eV in the
other.

\paragraph{Verification against a bifurcation of known size.}
Real data cannot confirm the model's uncertainty has the correct magnitude,
since the true conditional is never exposed: all eight inputs varied at
once, so no run holds seven fixed while sweeping the eighth through the
transition. Non-unique divertor solutions
have been reported in reduced neural surrogates
\citep{poels2023div1d, lore2024sparc},
so representing them matters, but verifying the spread's magnitude needs a
case where the answer is known. We construct one. The deterministic MLP surrogate
of \citet{dasbach2026solpsnn} is evaluated on inputs in which the perpendicular heat
diffusivity $\chi_\perp$ is drawn from one of two Gaussians,
$\mathcal{N}(0.3, 0.03)$ and $\mathcal{N}(1.6, 0.16)$, the particle diffusivity
fixed. Its outputs form a dataset of $14442$ points containing, by construction, a two-branch bifurcation with known within-branch noise. The same flow-matching architecture is then trained on it seeing only the remaining six inputs, so the branches appear as an unlabelled, irreducible ambiguity. This is not a deployment path; it exists to make the ground-truth uncertainty exactly known.

The split persists across the branch-coexistence region rather than at an
isolated point. Figure~\ref{fig:results}c compares the sampled
$\te^{\mathrm{OT}}$ distribution past the transition against exact branch
temperatures pooled from the $20$ nearest test points: because the construction
evaluates the same visible inputs under both $\chi_\perp$ populations, each
pooled point's two branch temperatures are known rather than approximated. The cold mode sits essentially on the cold branch; the
hot mode sits toward the upper part of the hot branch's spread, consistent
with the mild over-prediction above.

\section{Discussion and limitations}
\label{sec:discussion}

A curvilinear, fixed-topology edge mesh can be given to an ordinary
convolutional network as three image tensors without discarding adjacency or information, and where the steady state is ambiguous, a multi-modal predictive distribution is a more honest answer than its mean. Three limitations bound
this. We report no head-to-head accuracy comparison against
the deterministic baseline, so the contribution concerns what the surrogate
represents rather than a win on point accuracy.
Cold-core
solutions lie outside the fluid-neutral model's validity range, so target
and error there are indicative only. The representation is tied to one fixed-topology mesh, 
so extending it to other divertor geometries would require re-deriving strip boundaries and adjacencies by hand.

\clearpage
\bibliographystyle{plainnat}
\bibliography{refs}

\end{document}